\documentclass[runningheads]{llncs}
\usepackage[T1]{fontenc}
\usepackage{graphicx,verbatim}
\usepackage{amsmath}
\usepackage{amssymb}
\usepackage{arydshln}
\usepackage{bm}

\usepackage{lipsum} 
\usepackage[inline]{enumitem}
\usepackage{booktabs}
\usepackage[table]{xcolor} 
\definecolor{best}{RGB}{198,239,206}
\usepackage{xcolor}
\usepackage{pifont}
\usepackage{hyperref}
\usepackage{color}

\usepackage{adjustbox}
\usepackage{pgf}

\usepackage{subcaption}
\usepackage{makecell}

\begin{document}
%
\title{Mission Balance: Generating Under-represented Class Samples using Video Diffusion Models}
\titlerunning{Surgical video generation}
%
\author{Danush Kumar Venkatesh\inst{1,2}\and
Isabel Funke\inst{1} \and Micha Pfeiffer\inst{1} \and Fiona Kolbinger\inst{4,5,6} \and Hanna Maria Schmeiser{\inst6} \and Marius Distler\inst{6} \and J\"urgen Weitz\inst{6} \and Stefanie Speidel\inst{1,3}
}
\authorrunning{D. Venkatesh et al.}
%
\institute{\small{NCT/UCC Dresden, DKFZ Heidelberg, Faculty of Medicine \& University Hospital Carl Gustav Carus TU Dresden,HZDR Dresden, Germany \and Department of Translational Surgical Oncology, NCT/UCC Dresden, Faculty of Medicine \& University Hospital Carl Gustav Carus Germany \and The Centre for Tactile Internet with Human-in-the-Loop (CeTI), TUD Dresden, Germany \and Weldon School of Biomedical Engineering, Regenstrief Center for Healthcare Engineering (RCHE), Purdue University, USA \and Department of Biostatistics and Health Data Science, Richard M. Fairbanks School of Public Health, Indiana University, USA \and Department of Visceral, Thoracic and Vascular Surgery, University Hospital \& Faculty of Medicine Carl Gustav Carus, TUD Germany}
\\
\email{\{first,last name\}@nct-dresden.de,\{fkolbing\}@purdue.edu }}



\maketitle              
\begin{abstract}
Computer-assisted interventions can improve intraoperative guidance, particularly through deep learning methods that harness the spatiotemporal information in surgical videos. However, the severe data imbalance often found in surgical video datasets hinders the development of high-performing models. In this work, we aim to overcome the data imbalance by synthesizing surgical videos. We propose a unique two-stage, text-conditioned diffusion-based method to generate high-fidelity surgical videos for under-represented classes. Our approach conditions the generation process on text prompts and decouples spatial and temporal modeling by utilizing a 2D latent diffusion model to capture spatial content and then integrating temporal attention layers to ensure temporal consistency. Furthermore, we introduce a rejection sampling strategy to select the most suitable synthetic samples, effectively augmenting existing datasets to address class imbalance. We evaluate our method on two downstream tasks—surgical action recognition and intra-operative event prediction—demonstrating that incorporating synthetic videos from our approach substantially enhances model performance. We open-source our implementation at \url{https://gitlab.com/nct_tso_public/surgvgen}. 


\end{abstract}
\section{Introduction}
Computer-assisted intervention (CAI) aims to enhance surgical procedures by integrating advanced computational techniques~\cite{maier2022surgical}. 
The rapid growth of deep learning (DL) has further propelled CAI research, enabling applications such as surgical action recognition, phase recognition, and the identification of critical anatomical structures to offer context-aware guidance and decision support during surgery~\cite{maier2022surgical}. Notably, these applications can benefit from temporal context by analyzing surgical video sequences rather than individual frames. 

\begin{figure}[tb]
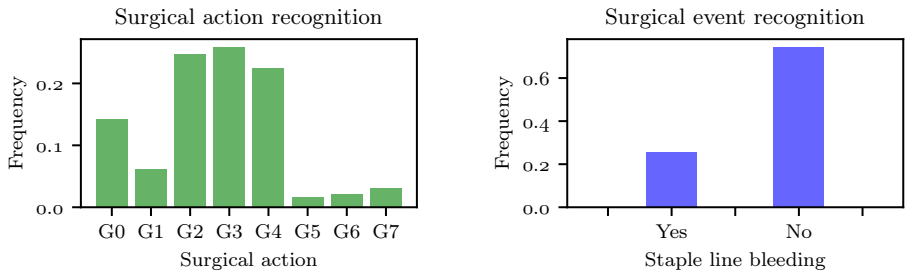

\centering
\begin{subfigure}{0.47\textwidth}
        \centering
        \begin{adjustbox}{width=\textwidth,keepaspectratio}
            \input{imbalance_sarrarp.pgf}%
        \end{adjustbox}
    \end{subfigure}
    \hfill
    \begin{subfigure}{0.47\textwidth}
        \centering
        \begin{adjustbox}{width=\textwidth,keepaspectratio}
            \input{imbalance_slb.pgf}%
        \end{adjustbox}
    \end{subfigure}
\caption{Data imbalance among classes in surgical datasets. Left: Frequency of different actions (G0-G7) in the SAR-RARP50~\cite{psychogyios2023sar} dataset. Right: Occurrence of staple line bleeding in distal pancreatectomy.} 
\label{imbal}
\end{figure}

However, DL methods require large volumes of real, diverse, and annotated surgical video data
~\cite{maier2022surgical}. Although the availability of public surgical datasets has increased, significant data imbalance remains a critical issue (see Fig.~\ref{imbal}), which causes DL methods to become biased toward the majority classes. While oversampling and data augmentation can mitigate this problem, they increase the frequency of under-represented samples but do not address the lack of diversity.

Synthesizing data with generative models offers a promising solution to the lack and imbalance of data~\cite{venkatesh2024exploring,frisch2023synthesising}. In particular, \emph{diffusion models (DMs)}~\cite{dhariwal2021diffusion} have been applied to generate high-quality surgical images conditioned on tissue texture and shape~\cite{VenkateshWACV,kaleta2023minimal}; however, these approaches lack temporal context—a critical factor for generating surgical videos. 
On the other hand, recent methods for surgical video synthesis \cite{li2024endora,iliash2024interactive,cho2024surgen} either (i) do not condition the generation on class labels, which is needed to control the generation of new samples specifically for underrepresented classes \cite{li2024endora}, (ii) require pre-existing instrument masks to control the synthesis of each video frame \cite{iliash2024interactive}, (iii) rely on large volumes of data (100–200K frames), which is typically not available for under-represented cases \cite{cho2024surgen}, or (iv) use spatio-temporal models with high computational demand \cite{li2024endora,cho2024surgen}.
In contrast, we propose to leverage a pre-trained 2D latent diffusion model -- \emph{Stable Diffusion (SD)}~\cite{rombach2022high} -- and extend it into a video diffusion model by adding temporal layers to separately model video dynamics. By conditioning the model on text prompts, we generate videos for specific classes. By separating the spatial and temporal modeling, the training and inference efficiency is improved. 
In particular, we start with a pre-trained SD model, which we finetune into a common text-conditioned spatial diffusion model for all classes. Based on that, we create the video diffusion model by freezing the spatial layers and adding temporal layers, which are trained with additional class label conditioning to learn class-specific dynamics for each under-represented class. 

In addition, we introduce a rejection sampling procedure to select the most suitable synthetic video samples for downstream surgical tasks.

We evaluate the abilities of \emph{SurV-Gen}, our proposed surgical video diffusion model, to synthesize videos of under-represented classes on two downstream tasks: \textcircled{1} surgical action recognition on the SAR-RARP50~\cite{psychogyios2023sar} dataset and \textcircled{2} video-level recognition of staple line bleeding (SLB) during pancreas transection on an in-house dataset. 
To the best of our knowledge, we are the first to tackle the data imbalance problem in surgical data science using diffusion models for video generation.

\section{Method}
\begin{figure}[tb]
\includegraphics[width=\textwidth]{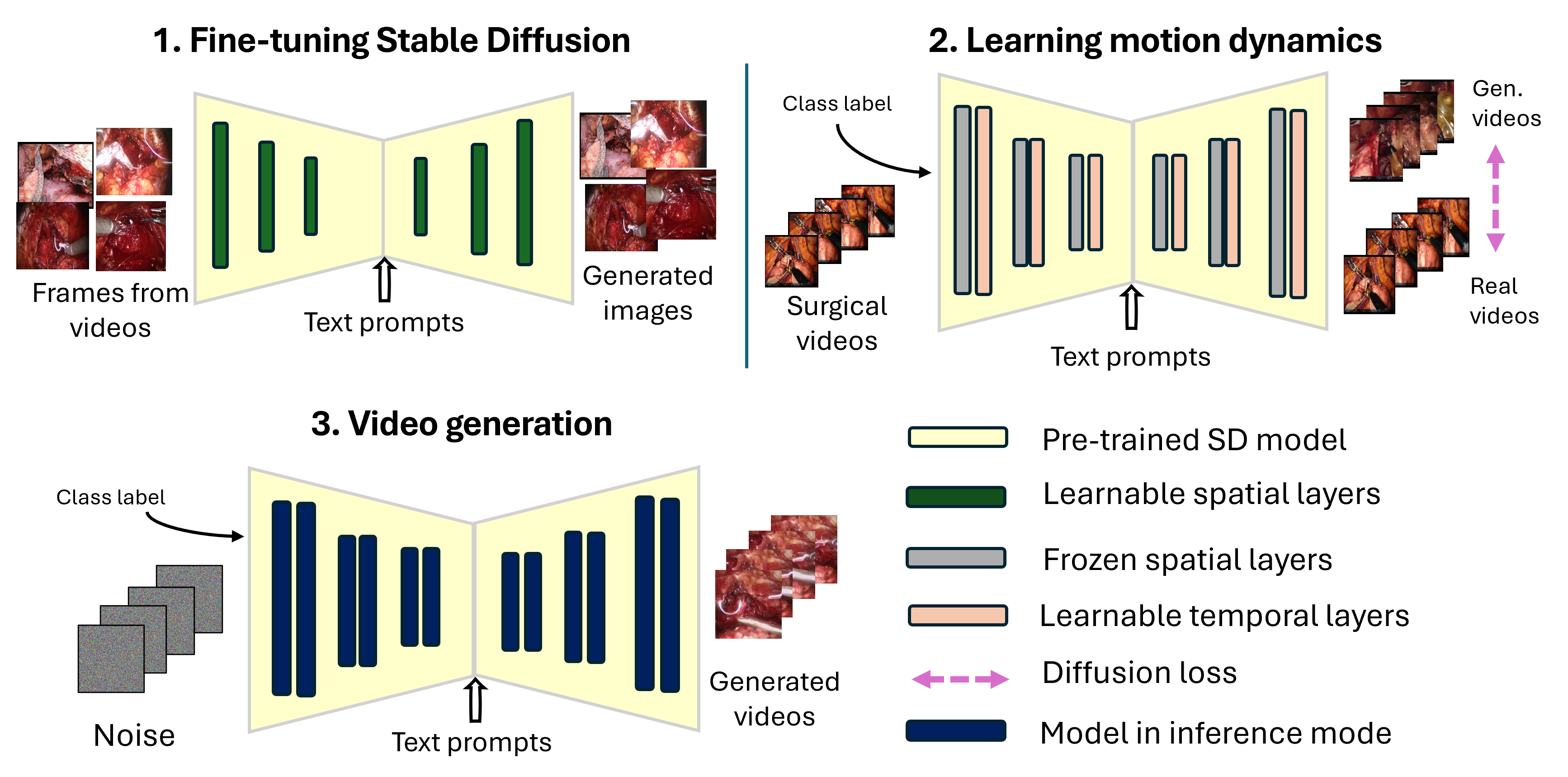}
\caption{Overview of the SurV-Gen method for surgical video generation. 
} 
\label{method}
\end{figure}
Our video diffusion method, SurV-Gen, is trained in two stages.
In stage~1, we fine-tune a pre-trained Stable Diffusion model on video frames paired with text prompts to capture the spatial content. In stage~2, with spatial layers frozen, temporal transformer layers are inserted and trained to model dynamics, with text conditioning applied throughout. During inference, SurV-Gen generates class-specific videos from text prompts, which are refined via a sample selection pipeline. An overview of the method is shown in Fig.~\ref{method}.

\subsection{Stage 1: Fine-tuning Stable Diffusion}
Stable Diffusion (SD)~\cite{rombach2022high} was opted as the base text-to-image model in this work because it is open-sourced and produces high-quality surgical images~\cite{MartyniakWACV,VenkateshWACV,kaleta2023minimal}. In SD, the diffusion process takes place in latent space by encoding the image~\(x_0\) to~\(z_0\) with an encoder $\mathcal{E}(x_0)$. During the forward diffusion process, \(z_0\)  is perturbed to \(z_t\) via 
\begin{equation}
z_t = \sqrt{1-\beta_t}\, z_{t-1} + \sqrt{\beta_t}\, \bm{\epsilon}_{t-1} = \sqrt{\bar{\alpha}_t}\, z_0 + \sqrt{1 - \bar{\alpha}_t}\, \bm{\epsilon}_0, \quad 1 \leq t \leq T, 
\end{equation}
where $\bm{\epsilon}_t \sim \mathcal{N}(0, I)$, $\alpha_t = 1 - \beta_t$, $\bar{\alpha}_t = \prod_{s=0}^t \alpha_s$, and the predefined $\beta_t$ controls the noise strength at each step $t$.
A denoising network \(\bm{\epsilon}_\theta(\cdot)\) is trained to reverse this process by predicting the noise that was added by minimizing loss $L$:
\begin{equation} \label{eq:1}
    L = \mathbb{E}_{\mathcal{E}(x_0),\, y,\, \bm{\epsilon} \sim \mathcal{N}(0,I),\, t} \left[ \left\|\bm{\epsilon} - \bm{\epsilon}_\theta\bigl(z_t, t, y\bigr) \right\|_2^2 \right], 
\end{equation}
where \(y\) is the text prompt associated with image~\(x_0\). \(\bm{\epsilon}_\theta(\cdot)\) is implemented as a U-Net~\cite{ronneberger2015u} and we perform a velocity estimate of \(\bm{\epsilon}\)~\cite{salimans2022progressive}. 
We fine-tune the SD model in this stage using images extracted from the videos together with text prompts that describe the corresponding class label.
\subsection{Stage 2: Learning motion dynamics}
In this stage, we extend the fine-tuned SD model to train directly on surgical video sequences consisting of 16 frames. We adopt the method proposed in recent works~\cite{ho2022video,blattmann2023align,guo2023animatediff} by incorporating temporal \emph{transformer} blocks~\cite{waswani2017attention} into the SD model, placing them after each spatial layer. 
Since this stage focuses solely on modeling temporal dynamics, the spatial layers are kept frozen.

Given a 5D video tensor \(v \in \mathbb{R}^{b \times c \times f \times h \times w}\), where \(b\) is the batch size, \(c\) and \(f\) are the number of channels and frames respectively, and \(h\) and \(w\) are the spatial dimensions,
the frozen spatial layers process $v$ frame-wise by reshaping to $(b f) \times c \times h \times w$. 
In the temporal layers, the sequences at each spatial location are processed independently by reshaping $v$ to $(b h w) \times f \times c$.
Here, sequence \(v_{\text{in}} \in \mathbb{R}^{(b h w) \times f \times c}\) is projected and -- after adding sinusoidal positional encoding -- processed using \emph{self-attention} via:
\begin{equation}
v_{\text{out}} = \text{Attention}(Q,K,V) = \text{Softmax}\left(\frac{QK^T}{\sqrt{c}}\right)V,
\end{equation}
with \(Q = v_{\text{in}} W_Q\), \(K = v_{\text{in}} W_K\), and \(V = v_{\text{in}} W_V\) being query, key and value vectors. In this way, each generated frame incorporates information from other frames in the video clip, capturing the motion dynamics over time. The self-attention is followed by a multi-layer perceptron. Following~\cite{zhang2023adding}, we initialize the output projection layers of the temporal blocks to zero and include a residual connection. In addition to text prompts, we add an embedding of the class label during training to serve as an additional conditioning signal and train with the same loss as in Eq.~\ref{eq:1}. 

\subsection{Selection of generated data}
A common approach for leveraging synthetically generated data in downstream tasks is to combine it with real datasets to improve performance~\cite{frisch2023synthesising,li2024endora}. However, prior works have shown that adding synthetic data can sometimes have adverse effects~\cite{azizi2023synthetic,alaa2022faithful}. To address this, we introduce a \emph{rejection sampling} (RS) procedure to select the most suitable synthetic videos from the pool of generated candidates. Specifically, we train a discriminative model to predict class labels on the available real datasets and then use this model to evaluate the synthetic videos. A synthetic video that was conditionally generated for class label $l$ is only retained if the model's top-$k$ predictions contain $l$.

\section{Evaluation}

\subsection{Experiments}
\subsubsection{Downstream tasks}
For downstream evaluation, we use two challenging tasks on small-scale intraoperative video datasets with imbalanced class distribution.

\emph{Action recognition} (Task\,\textcircled{1}):
The task is to recognize which surgical action is performed at any time $t$ in a video. 
Here, we use the SAR-RARP50\footnote{Train set: \url{https://doi.org/10.5522/04/24932529.v1}, test set: \url{https://doi.org/10.5522/04/24932499.v1}, license: CC BY-NC-SA 4.0} dataset~\cite{psychogyios2023sar}, which consists of 50~videos of robot-assisted suturing with an average duration of ca. 5\,min. We split them into 35, 5, and 10 videos for training, validation, and testing. Seven different actions plus a background class are defined, including the under-represented actions ``Picking-up the needle'' (G1), ``Cutting the suture'' (G6), and ``Returning/dropping the needle'' (G7) (see Fig.~\ref{imbal}). Action G5 (``Tying a knot'') is not considered further because it occurs only in one test video.
As DL model, we train X3D~\cite{feichtenhofer2020x3d}, a 3D\,CNN for video recognition, to recognize the action at time~$t$ by classifying the 16-frame video clip that is centered around time~$t$. To measure model performance, we compute the average video-wise Jaccard index for each action.  

\begin{table}[tb]
\caption{Image-level quality comparison of generated video frames for SLB recognition task after RS. Inf. time denotes the time for generating a video.} \label{tab:quality}
  \begin{center}
  {\small{
    \resizebox{\linewidth}{!}{
    \begin{tabular}{lcccccc}
    \toprule
    Method & \makecell[cc]{Trainable\\params} & Inf. time & CFID ($\downarrow$) & Density($\uparrow$) & Coverage($\uparrow$) & CMMD($\downarrow$)  \\
\midrule
LVDM~\cite{he2022latent} & $548$M & $43.3$s & $185.45_{\pm4.30}$ & $0.20_{\pm0.007}$ &  $0.40_{\pm0.006}$ & $3.98_{\pm0.10}$ \\
Endora~\cite{li2024endora} & $675$M & $15.3$s & $\underline{117.72_{\pm2.93}}$ & $\underline{0.81_{\pm0.004}}$ &  $\mathbf{0.71_{\pm0.001}}$ & $\underline{2.78_{\pm0.001}}$ \\
SurV-Gen & $435$ M & $6.55$s & $\mathbf{108.30_{\pm1.22}}$ & $\mathbf{0.90_{\pm0.002}}$ &  $\underline{0.65_{\pm0.003}}$ & $\mathbf{2.25_{\pm0.01}}$ \\
    \bottomrule
\end{tabular}
}}}
\end{center}
\end{table}

\emph{SLB recognition} (Task\,\textcircled{2}): 
The task is to recognize whether or not staple line bleeding (SLB) occurs in a video of the pancreas transection phase during distal pancreatectomy.
During pancreatic stapler transection, SLB represents a visually recognizable event that has been associated with postoperative pancreatic fistula \cite{zimmitti2021investigation}.
We use an internal dataset of 39 videos, each 3\,min long on average, where SLB occurred in 10 cases. The analysis of these deidentified data was approved by the local institutional review board, and informed patient consent was waived. 
We split the data into 25, 4, and 10 videos for training, validation, and testing.
As DL model, we train a ResNet-LSTM on the videos, where the ResNet~\cite{he2016deep} extracts visual features from individual video frames and the LSTM~\cite{hochreiter1997long} aggregates these features over time. A linear classifier recognizes the occurrence of SLB based on the final LSTM state. To measure model performance, we compute balanced accuracy and $F_1$ score.

\begin{figure}[tbp]
\includegraphics[width=\textwidth]{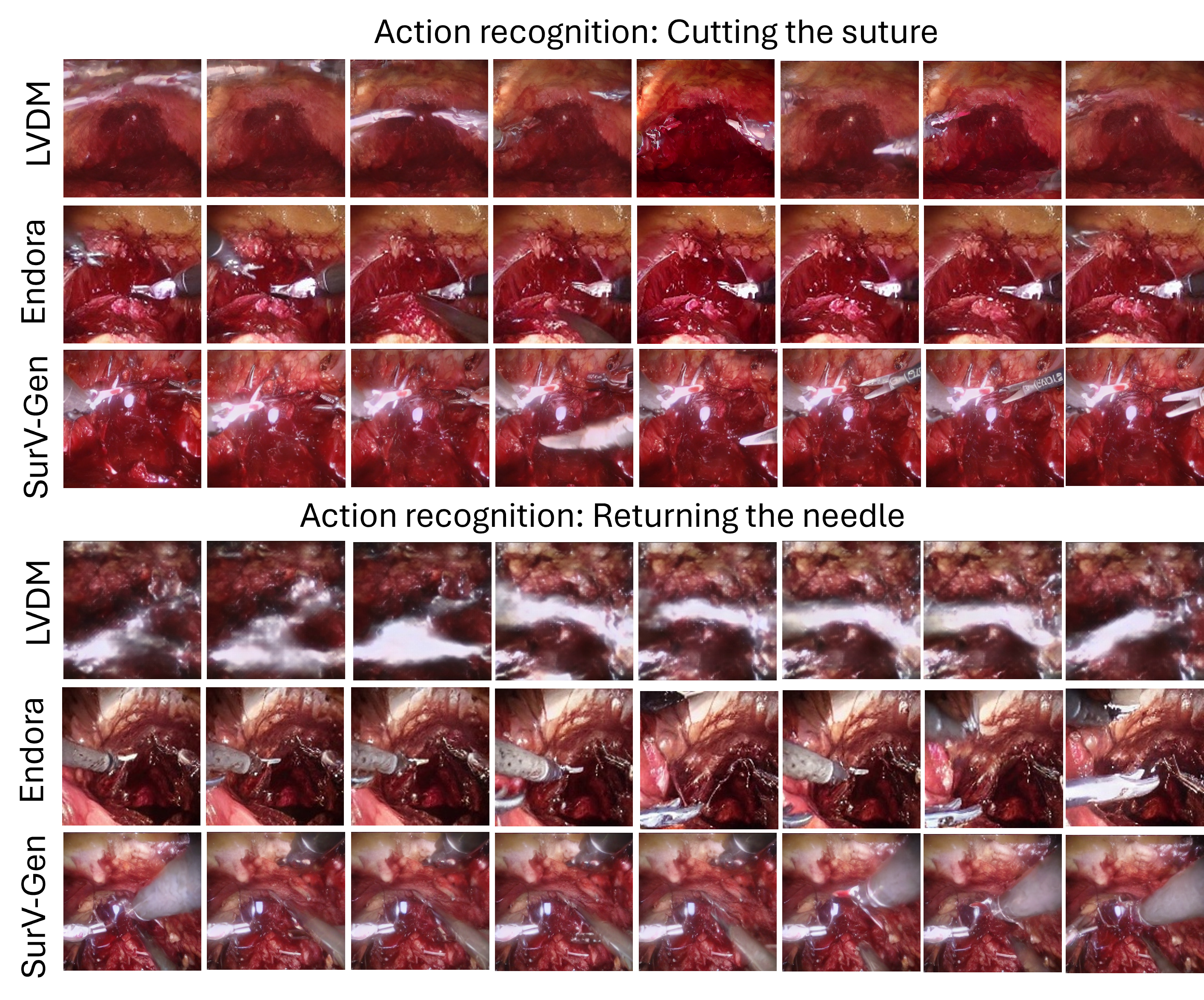}
 \caption{Qualitative comparison of generated video frames. In our approach (SurV-Gen), the scissors are clearly visible during suture cutting (row 3), whereas in other methods, the scissors appear only partially. Similarly, during the needle return, the tools are consistently generated (row 6).} 
\label{comp}
\end{figure}

Notably, we want to investigate the impact of adding synthetic training data and therefore chose solid baselines for the downstream tasks instead of more sophisticated models. We repeat all experiments on the downstream tasks three times and average the evaluation metrics over these runs.

\textbf{Baselines} We selected Endora~\cite{li2024endora} and LVDM~\cite{he2022latent} as the video diffusion baselines. Endora is a fully transformer-based diffusion model that jointly models spatio-temporal components whereas LVDM uses a 3D U-Net architecture for a 3D latent video diffusion approach. These methods were chosen as strong baselines due to their open-source implementations, which allowed us to re-implement them effectively. 

\textbf{Video generation} 
On task\,\textcircled{1}, we trained a single SD model for all the selected classes (G1, G6, G7), after which separate spatio-temporal models were trained for the three classes using label conditioning. Text prompts were constructed based on the class label as ``action recognition task of <gesture>'' (task\,\textcircled{1}) and ``a complication of staple line bleeding'' (task\,\textcircled{2}) for both training and inference. We train the diffusion models on the respective training sets of the downstream tasks and generate videos of  resolution of $256$×$256$ pixels for SurV-Gen and LVDM, and $128$×$128$ pixels for the Endora model. 

\textbf{Downstream evaluation}
For evaluating any of the video generation methods, we generated videos for each under-represented class and performed rejection sampling (RS) using a lightweight ResNet3D classifier \cite{hara2018can} with thresholds of \(k=3\) (task\,\textcircled{1}) and \(k=1\) (task\,\textcircled{2}). 
With SurV-Gen, we further conducted an ablation study without rejection sampling.


\subsection{Results and discussion}
We evaluate the generated videos for both image quality and -- by measuring their benefit for downstream tasks -- realism, diversity, and temporal consistency. 

\textbf{Image quality} We evaluate the fidelity and diversity of the generated frames using (i) the CFID metric~\cite{parmar2022aliased} to quantify realism, (ii) the CMMD score~\cite{jayasumana2024cmmd} to quantify unbiased image quality, which is particularly effective for smaller datasets, and (iii) density and coverage  metrics~\cite{naeem2020reliable} to assess image quality and diversity.
 
The results in Tab.~\ref{tab:quality} indicate that our SurV-Gen model synthesizes high-quality surgical video frames compared to baseline methods. This finding is further supported by the qualitative results in Fig.~\ref{comp}, where SurV-Gen effectively generates surgical instruments, such as scissors and graspers, corresponding to specific gestures. In contrast, the baseline models either omitted these tools or produced poorly defined scenes, which we attribute to their dependence on larger datasets for learning temporal dynamics. Moreover, the SurV-Gen model features fewer trainable parameters and requires less inference time than the baselines. Additional generated videos are provided as supplementary material.

\begin{table}[h]
\caption{Results on task\,\textcircled{1}. We report the Jaccard index for each class, including \colorbox{best}{under-represented} classes. (RS: rejection sampling)
}
\label{tab:ph_comp}
\centering
\begin{subtable}{0.8\linewidth}
\raggedright
\begin{tabular}{lccccc}
\toprule
Training data & ~~RS~~ & G0 & \cellcolor{best}G1 & G2 & G3  \\
\midrule
Only Real & - &  $0.54_{\pm0.002}$ & $0.31_{\pm0.004}$ & $0.62_{\pm0.004}$ & $0.75_{\pm0.03}$ \\
\midrule
Real + LVDM & \checkmark & $0.47_{\pm0.01}$ & $0.30_{\pm0.03}$ & $0.60_{\pm0.01}$ & $\mathbf{0.77_{\pm0.009}}$ \\
Real + Endora & \checkmark & $\underline{0.54_{\pm0.03}}$ & $\mathbf{0.34_{\pm0.06}}$ & $\underline{0.61_{\pm0.01}}$ & $\mathbf{0.77_{\pm0.01}}$  \\
Real + SurV-Gen & \checkmark & $\mathbf{0.55_{\pm0.01}}$ & $\underline{0.31_{\pm0.01}}$ & $\mathbf{0.62_{\pm0.005}}$ & $\underline{0.76_{\pm0.001}}$ \\
\midrule
Real + SurV-Gen & \ding{55} & $0.52_{\pm0.10}$ & $0.25_{\pm0.05}$ & $0.60_{\pm0.01}$ & $0.75_{\pm0.002}$  \\
    \bottomrule
\end{tabular}
\end{subtable}
\hfill
\begin{subtable}{0.8\linewidth}
\raggedleft
\begin{tabular}{cccc|c}
\toprule
G4 & G5$^{*}$ & \cellcolor{best} G6 & \cellcolor{best} G7 & Avg. \\
\midrule
$0.58_{\pm0.02}$ & $0.19_{\pm0.10}$ & $0.11_{\pm0.03}$ & $0.23_{\pm0.06}$ & $0.44_{\pm0.11}$ \\
\midrule
$0.59_{\pm0.01}$ & $\underline{0.25_{\pm0.13}}$ & $0.20_{\pm0.01}$ & $0.23_{\pm0.04}$ & $0.44_{\pm0.07}$ \\
$\underline{0.60_{\pm0.01}}$ & $\mathbf{0.26_{\pm0.13}}$ & $\underline{0.20_{\pm0.01}}$ & $\underline{0.31_{\pm0.02}}$ & $\underline{0.48_{\pm0.09}}$ \\
$\mathbf{0.63_{\pm0.02}}$ & $0.18_{\pm0.14}$ & $\mathbf{0.23_{\pm0.03}}$ & $\mathbf{0.37_{\pm0.01}}$ & $\mathbf{0.50_{\pm0.10}}$ \\
\midrule
$0.60_{\pm0.01}$ & $0.10_{\pm0.09}$ & $0.18_{\pm0.009}$ & $0.24_{\pm0.003}$ & $0.44_{\pm0.14}$\\
    \bottomrule
\multicolumn{5}{r}{$^{*}$G5 occurs only in one test video.} 
\end{tabular}
\end{subtable}
\end{table}

\begin{table}[tb]
\caption{Results on SLB recognition (task\,\textcircled{2}). (RS: rejection sampling)}\label{tab:slb_task}
\begin{center}
\begin{tabular}{lccc}
\toprule
    Training data & ~RS~ & Bal. Acc ($\uparrow$)~ & $F_1$ ($\uparrow$)\\
\midrule
Only Real & - &  $0.74_{\pm0.08}$ & $0.71_{\pm0.01}$\\
\midrule
Real + LVDM & \checkmark &  $0.72_{\pm0.10}$ & $0.69_{\pm0.03}$\\Real + Endora & \checkmark  &  $\underline{0.76_{\pm0.01}}$ & $\underline{0.75_{\pm0.10}}$\\
Real + SurV-Gen & \checkmark &  $\mathbf{0.81_{\pm0.05}}$ & $\mathbf{0.78_{\pm0.04}}$\\
\midrule
Real + SurV-Gen & \ding{55} &   $0.74_{\pm0.08}$ & $0.70_{\pm0.10}$\\
    \bottomrule
\end{tabular}
\end{center}
\end{table}


\textbf{Downstream evaluation} Tab.~\ref{tab:ph_comp} presents the results for task\,\textcircled{1}. Extending the training set with synthetically generated videos enhanced performance across all baseline models for most classes. Notably, our SurV-Gen model with rejection sampling achieved improvements of $12$ and $14$ points for classes G6 and G7, respectively, relative to using only real videos. These results underscore the high quality of the synthesized samples and the effectiveness of the rejection sampling strategy. A modest improvement was observed for gesture G1 only with the Endora model.  Further exploration of model architectures with the addition of more synthetic samples can improve performance, which we leave for future work.

The results for SLB recognition is shown in Tab.~\ref{tab:slb_task}. The addition of synthetic videos to the real training data improved performance for both the SurV-Gen and Endora models compared to relying solely on real videos. Conversely, the LVDM model did not show any performance improvements. 

The ablation on rejection sampling underscores the critical role of synthetic sample selection on both downstream tasks. For example on task\,\textcircled{2}, SurV-Gen with RS achieves a 7-point boost whereas the gains without rejection sampling are only modest. 

\section{Conclusion}
In this work, we introduce SurV-Gen, a light-weight two-stage diffusion framework synthesizing high-fidelity surgical videos for underrepresented classes, thereby addressing the data imbalance in surgical datasets. By separating spatial and temporal modeling,  our approach can directly reuse pre-trained weights of large-scale image diffusion models, which efficiently helps to learn video generation with limited examples. For the first time, we show that synthesizing additional training videos for under-represented classes helps to improve the performance of video recognition models on two challenging surgical downstream tasks with data imbalance. Here, selecting the most suitable synthetic videos using rejection sampling proved to be a crucial step. Future work may extend this framework to model additional under-represented classes, such as surgical tools, by incorporating conditional signals, e.g., segmentation maps, from surgical simulations. Furthermore, a more detailed analysis of the contribution of individual synthetic samples to downstream performance could provide valuable insights for further improving video generation.

%
%

    

\begin{credits}
\subsubsection{\ackname} This work is partly supported by BMBF (Federal Ministry of Education and Research) in DAAD project 57616814 (\href{https://secai.org/}{SECAI, School of Embedded Composite AI}). This project was also partially funded by the German Research Foundation (DFG, Deutsche Forschungsgemeinschaft) as part of Germany’s Excellence Strategy – EXC 2050/1 –Project ID 390696704 – Cluster of Excellence “Centre for Tactile Internet with Human-in-the-Loop” (CeTI) of Technische Universit\"at Dresden.

\subsubsection{\discintname}
The authors have no competing interests to declare that are
relevant to the content of this article.
\end{credits}


%
%
%
\bibliographystyle{splncs04}
\bibliography{main}

\end{document}